\documentclass[]{interact}

\usepackage{epstopdf}% To incorporate .eps illustrations using PDFLaTeX, etc.
\usepackage{subcaption}
\usepackage{makecell}
\usepackage{natbib}  % Enables author-year citation
\usepackage{float} % In your preamble

\bibpunct[, ]{(}{)}{;}{a}{}{,}% Citation support using natbib.sty
\theoremstyle{plain}% Theorem-like structures provided by amsthm.sty

\theoremstyle{definition}

\theoremstyle{remark}

\begin{document}

\articletype{ARTICLE TEMPLATE}

\title{Towards Scalable and Generalizable Earth Observation Data Mining via Foundation Model Composition}

\author{
\name{Man Duc Chuc}
\affil{University of Engineering and Technology, Vietnam National University, Hanoi, Vienam}
}

\maketitle

\begin{abstract}
Foundation models are rapidly transforming Earth Observation data mining by enabling generalizable and scalable solutions 
for key tasks such as scene classification and semantic segmentation. While most efforts in the geospatial domain 
have focused on developing large models trained from scratch using massive Earth Observation datasets, 
an alternative strategy that remains underexplored is the reuse and combination of existing pretrained models. 
In this study, we investigate whether foundation models pretrained on remote sensing and general vision datasets 
can be effectively combined to improve performance across a diverse set of key Earth Observation tasks. 
Using the GEO-Bench benchmark, we evaluate several prominent models, including Prithvi, Hiera, and DOFA, 
on eleven datasets covering a range of spatial resolutions, sensor modalities, and task types. 
The results show that feature-level ensembling of smaller pretrained models can match or exceed the performance 
of much larger models, while requiring less training time and computational resources. 
Moreover, the study highlights the potential of applying knowledge distillation 
to transfer the strengths of ensembles into more compact models, offering a practical path 
for deploying foundation models in real-world Earth Observation applications.
\end{abstract}

\begin{keywords}
foundation model; Prithvi; Hiera; SAM, DOFA; ensemble; knowledge distillation; Earth Observation
\end{keywords}

\section{Introduction}

Foundation models are transforming Earth Observation (EO) data analysis. In the past few years, 
many foundation models have been introduced to learn generalizable representations from massive EO data, 
achieving remarkable performance on various tasks including scene classification and semantic segmentation \cite{Cong2022,Jakubik,Hong2024,Xiong,Szwarcman2024,Duc2025}. 
These models are pretrained on vast unlabeled remote sensing datasets via self-supervised learning and then finetuned for 
downstream tasks. An example is Prithvi-EO-2.0, a transformer-based foundation model developed 
by IBM and NASA \cite{Szwarcman2024}. Prithvi-EO-2.0 was pretrained on over 4.2 million global time series samples of multispectral images from the Harmonized 
Landsat–Sentinel-2 (HLS) data archive. The dataset encompasses multi-year, 30m resolution images over 
the contiguous United States. The model was pretrained using masked autoencoder framework to learn rich spatiotemporal features. 
The resulting models can be finetuned to diverse tasks such as flood mapping, 
wildfire burn scar segmentation, multi-temporal crop type mapping. 
Finetuning Prithvi-EO-2.0 was found to have faster convergence as training from random initialization, 
and the pretrained model was reported to outperform state-of-the-art (SOTA) methods in many tasks.
This represents a shift from task-specific models toward pretrained models that accelerates downstream learning. 

These improvements underscore the significant impact that large-scale pretraining and model scaling can have on performance in key remote sensing tasks. 
The emerging trend of training large models from scratch on remote sensing data has enabled the development of novel architectures specifically tailored 
to the EO domain. For example, researchers are moving beyond single-modality backbones toward unified multi-sensor models. 
An example is the Dynamic One-For-All (DOFA) architecture, which employs a dynamic hypernetwork that conditionally adjusts its weights 
based on the input sensor modality \cite{Xiong}. This allows a single backbone to process data from five different sensor types, 
each with distinct spectral characteristics. Trained jointly on a combination of five satellite data sources, 
DOFA learns a highly adaptable representation that generalizes across 14 diverse EO tasks, even for data 
from previously unseen sensors. These results highlight the growing momentum in remote sensing 
toward general-purpose foundation models trained on petabyte-scale archives, which can be finetuned for high-impact applications 
ranging from urban mapping to disaster response.

Despite recent progress, the predominant focus in the EO domain remains 
on developing dedicated remote sensing foundation models, such as Prithvi and DOFA, trained 
exclusively on domain-specific datasets. However, an alternative paradigm with 
significant potential remains underexplored. This involves leveraging existing pretrained models 
from within the EO domain and even from broader computer vision domains, rather than training new models from scratch. 
Techniques widely adopted in general AI such as model ensembling \cite{Jiang2023,Huang2024} and knowledge distillation \cite{Hinton2015,Gou2021}, 
where information is transferred from one or more large teacher models to a smaller student model, 
are still relatively uncommon in EO research.

In mainstream computer vision, large-scale foundation models like Meta’s Segment Anything Model (SAM) 
have demonstrated remarkable zero-shot capabilities, producing accurate segmentation masks for 
virtually any object with minimal prompting \cite{Kirillov}. Trained on an unprecedented dataset 
of approximately 11 million natural images with over 1 billion segmentation masks, 
SAM has emerged as a highly general model capable of capturing fine-grained visual features. 
Although SAM was not originally developed for remote sensing applications, 
there has been increasing interest in its use for EO tasks such as land cover classification.
Early investigations suggest that SAM can delineate features like buildings, roads, and water bodies 
in satellite imagery without task-specific training. The accuracy can be further increased by finetuning the model \cite{Osco2023,Ren2024}. 
These efforts highlight the potential of reusing large pretrained vision models in remote sensing, 
offering an efficient alternative to training models with hundreds of millions of parameters from scratch.

The limited number of studies exploring this cross-domain adaptation, such as applying 
SAM to EO data have already reported promising results, including improved zero-shot 
classification and faster annotation. 
These findings suggest that the EO community could significantly benefit from the broader AI community’s 
investments in foundation models. Bridging the gap between domain-specific and general-purpose 
models will likely be a promising direction for future research.
In this research, we investigate whether existing pretrained foundation models, specifically 
those from general computer vision and the EO domain, can be effectively 
combined to improve performance on a variety of remote sensing tasks, without the need to train larger models from scratch. 
This approach has the potential to inform future strategies for model ensembling and knowledge distillation, 
enabling the construction of new SOTA foundation models with significantly lower computational cost. 
In doing so, it contributes to advancing the broader development of foundation models in remote sensing and beyond.

\section{Methodology}

\subsection{Datasets}

\begin{figure}[h]
  \centering
  
  \begin{subfigure}[b]{\textwidth}
    \centering
    \includegraphics[width=\linewidth]{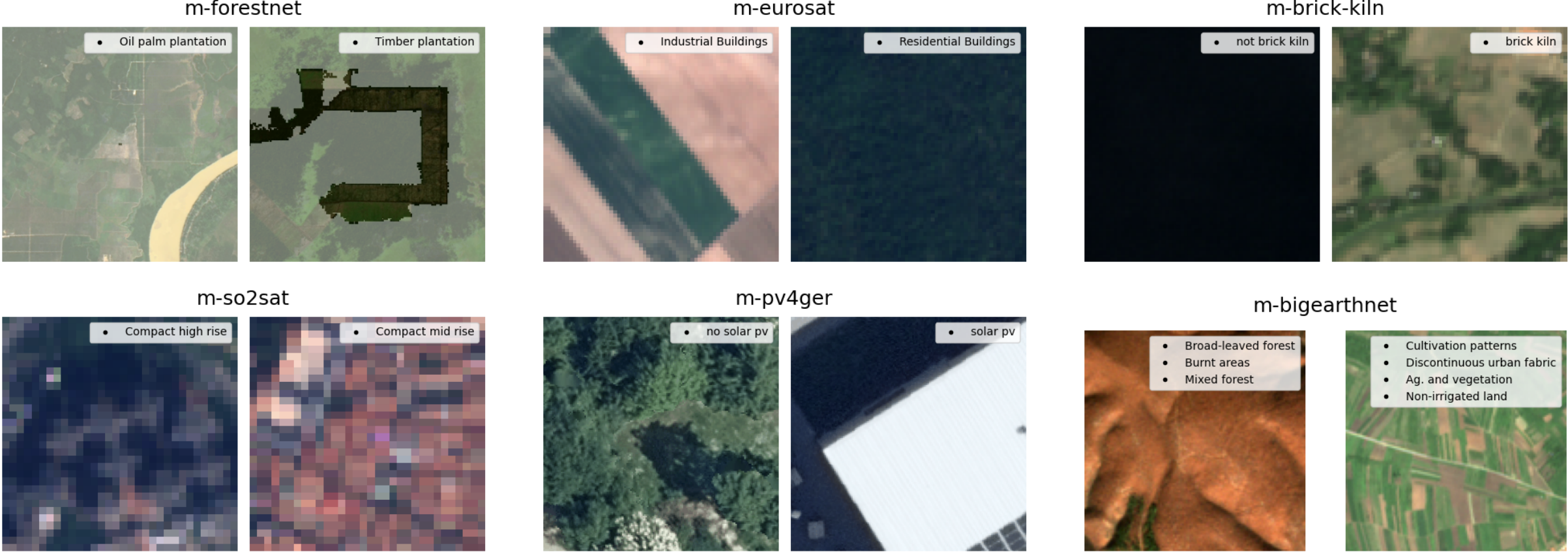}  % example image
    \caption{}
    % \label{fig:sub1}
  \end{subfigure}
  
  \vspace{0.5cm}

  \begin{subfigure}[b]{\textwidth}
    \centering
    \includegraphics[width=\linewidth]{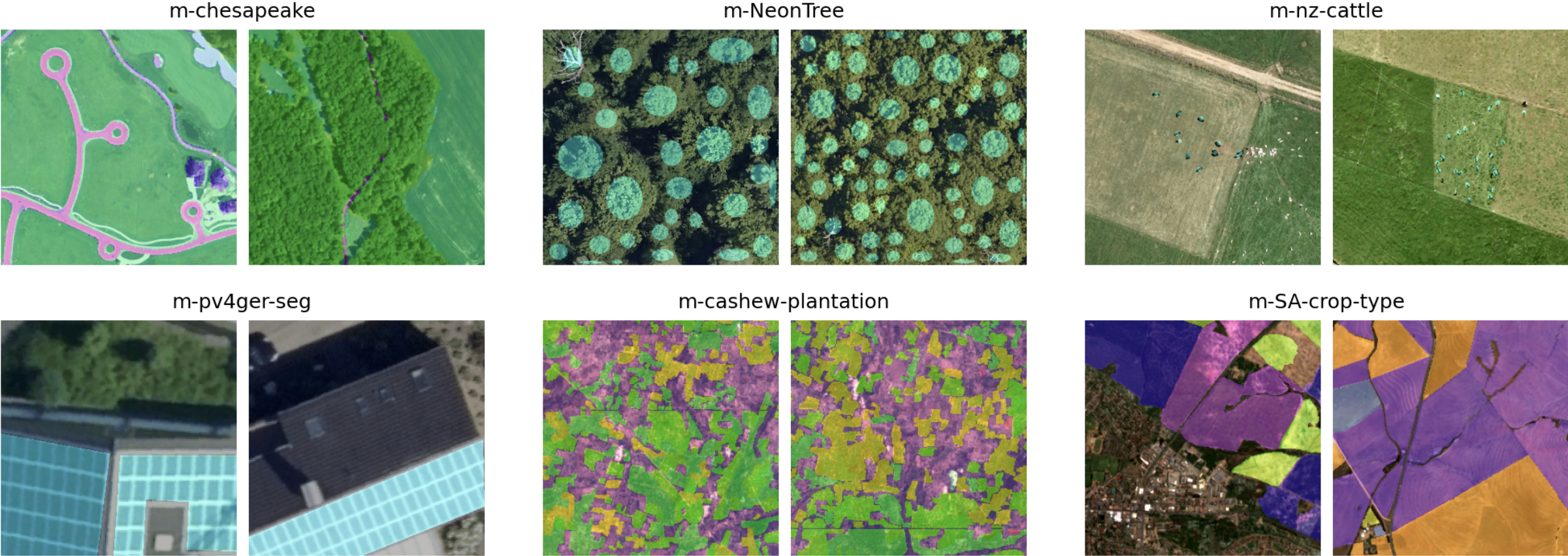}  % example image
    \caption{}
    % \label{fig:sub2}
  \end{subfigure}

  \caption{Representative examples of (a) classification and (b) segmentation tasks in the GEO-Bench datasets.}
  \label{fig:datasets}
\end{figure}

We used GEO-Bench, one of the most widely adopted and rigorous benchmarking frameworks 
for evaluating Earth Observation (EO) foundation models \cite{Lacoste2023}. 
The framework includes six scene classification datasets and six semantic segmentation datasets, 
spanning a range of spatial resolutions, dataset sizes, and application domains 
(see Table~\ref{tab:geo-bench} and Figure~\ref{fig:datasets}). GEO-Bench supports a variety of downstream tasks, 
including single-label classification, multi-label classification, and semantic segmentation (i.e., land cover classification).
However, in our experiments, we were unable to obtain the label file for the m-eurosat dataset and were therefore unable to include it in our evaluation.
For implementation, we adopted the TerraTorch library \cite{Gomes2025}, an open-source Python framework 
that allows users to easily modify or extend the existing codebase. TerraTorch comes 
with several pre-integrated backbones, including both versions of Prithvi, DOFA, ViT, 
and various CNN-based pretrained models. It also supports multiple datasets, 
including those from GEO-Bench. Training a model in TerraTorch is straightforward. 
Users simply define a configuration file specifying the backbone, dataset, training parameters, and other relevant settings.
In this study, in addition to the existing backbones, we integrated two new backbones (\textit{Hiera\_200M} and \textit{Hiera\_Prithvi\_500M}) 
into the TerraTorch library. These will be described in detail in the following sections.

\begin{table}[ht]
	\centering
	\caption{Characteristics of GEO-Bench datasets \cite{Lacoste2023}.}
	\label{tab:geo-bench}
	\resizebox{\textwidth}{!}{%
	\begin{tabular}{lcccccccc}
		\toprule
		\multicolumn{9}{c}{\textbf{Classification}} \\
		\midrule
		\textbf{Name} & \textbf{Image Size} & \textbf{\# Classes} & \textbf{Train} & \textbf{Val} & \textbf{Test} & \textbf{\# Bands} & \textbf{RGB res} & \textbf{Sensors} \\
		\midrule
		m-bigearthnet         & 120 $\times$ 120 & 43 & 20,000 & 1,000 & 1,000 & 12 & 10.0 & Sentinel-2 \\
		m-so2sat              & 32 $\times$ 32  & 17 & 19,992 & 986 & 986 & 18 & 10.0 & \makecell{Sentinel-2 \\ + Sentinel-1} \\
		m-brick-kiln          & 64 $\times$ 64  & 2  & 15,063 & 999 & 999 & 13 & 10.0 & Sentinel-2 \\
		m-forestnet           & 332 $\times$ 332 & 12 & 6,464 & 989 & 993 & 6 & 15.0 & Landsat-8 \\
		m-eurosat             & 64 $\times$ 64  & 10 & 2,000 & 1,000 & 1,000 & 13 & 10.0 & Sentinel-2 \\
		m-pv4ger              & 320 $\times$ 320 & 2 & 11,814 & 999 & 999 & 3 & 0.1 & RGB \\
		\midrule
		\multicolumn{9}{c}{\textbf{Segmentation}} \\
		\midrule
		\textbf{Name} & \textbf{Image Size} & \textbf{\# Classes} & \textbf{Train} & \textbf{Val} & \textbf{Test} & \textbf{\# Bands} & \textbf{RGB res} & \textbf{Sensors} \\
		\midrule
		m-pv4ger-seg          & 320 $\times$ 320 & 2  & 3,000 & 403 & 403 & 3 & 0.1 & RGB \\
		m-chesapeake-landcover & 256 $\times$ 256 & 7  & 3,000 & 1,000 & 1,000 & 4 & 1.0 & RGBN \\
		m-cashew-plantation   & 256 $\times$ 256 & 7  & 1,350 & 400 & 50 & 13 & 10.0 & Sentinel-2 \\
		m-SA-crop-type        & 256 $\times$ 256 & 10 & 3,000 & 1,000 & 1,000 & 13 & 10.0 & Sentinel-2 \\
		m-nz-cattle           & 500 $\times$ 500 & 2  & 524 & 66 & 65 & 3 & 0.1 & RGB \\
		m-NeonTree            & 400 $\times$ 400 & 2  & 270 & 94 & 93 & 5 & 0.1 & \makecell{RGB \\ + Hyperspectral \\ + Elevation} \\
		\bottomrule
	\end{tabular}%
	}
\end{table}

\subsection{Foundation models}
\subsubsection{Prithvi}

The Prithvi models represent a major advancement in general-purpose geospatial artificial intelligence. 
Released in December 2024, the second-generation model, Prithvi-EO-2.0 \cite{Szwarcman2024}, improves upon its predecessor, 
Prithvi-EO-1.0 \cite{Jakubik}, by introducing architectural enhancements and leveraging a significantly larger, 
globally distributed training dataset. Built on the Vision Transformer architecture, Prithvi-EO-2.0 incorporates temporal and location embeddings, 
enabling it to more effectively capture the complex spatiotemporal patterns found in EO data. These improvements have led to notable gains in 
performance across a wide range of EO tasks. The model was pretrained on a large-scale dataset of 4.2 million time series samples 
from HLS archive, which provides imagery at a 30-meter resolution. 
Two primary model sizes were released: Prithvi 300M ($\sim300$ million parameters) and Prithvi 600M ($\sim630$ million parameters), 
with each available in both a base version and an enhanced version that includes temporal and location embeddings. 
Benchmarking results show that the 600M variants often outperform the smaller models on tasks such as GEO-Bench, 
Wildfire Scar Mapping, and Burn Intensity Mapping. Training these models required substantial computational resources: 
the 300M model was trained on 80 GPUs for approximately 21,000 GPU-hours, while the 600M model required 240 GPUs and roughly 58,000 GPU-hours. 
In this study, we primarily used the variants augmented with temporal and location embeddings, 
particularly for downstream tasks that included temporal and/or spatial metadata. 
These models are referred to as \textit{Prithvi\_300M} and \textit{Prithvi\_600M}, respectively.

\subsubsection{Segment Anything Model}

The Segment Anything Model, introduced by Meta AI in April 2023, introduced promptable image segmentation \cite{Kirillov}. 
Its architecture consists of an image encoder; a prompt encoder designed to handle point, box, mask prompts and some initial text prompts; 
and a lightweight mask decoder, enabling fast and interactive segmentation. Trained on the large-scale SA-1B dataset, 
which contains over 1 billion masks across 11 million images, SAM exhibited impressive zero-shot generalization, 
capable of segmenting previously unseen objects without additional fine-tuning. Building on this success, 
Meta released SAM 2 in August 2024, extending segmentation capabilities to video \cite{Ravi}. 
SAM 2 adopts the Hiera architecture \cite{Ryali2023} as its image encoder and introduces a memory bank module that enables real-time segmentation of both images and videos. 
This addition allows the model to retain information across frames, effectively managing occlusions and object reappearances. 
To support training, Meta developed the SA-V dataset, one of the largest video segmentation datasets to date, containing 35.5 million masks 
across 50.9 thousand videos. Evaluation results show that SAM 2 significantly improves video segmentation accuracy 
while reducing the number of required user interactions by a factor of three compared to earlier approaches. 
In image segmentation, it outperforms the original SAM, achieving six times faster inference and higher accuracy, 
thanks largely to the efficiency of the Hiera encoder. Training SAM 2 demanded considerable computational resources, 
reportedly involving 256 A100 GPUs over 108 hours. In our work, we utilized the large version of SAM 2, 
specifically extracting the Hiera image encoder ($\sim$212M parameters), referred to as \textit{Hiera\_200M}, 
which constitutes approximately 95\% of the full SAM 2 model ($\sim$224M parameters), 
and used it as a backbone in the same manner as other backbone architectures.

\subsubsection{DOFA}

The DOFA (Dynamic One-For-All) model, introduced in June 2024, is a multimodal EO foundation model inspired 
by the biological concept of neural plasticity \cite{Xiong}. It is designed to enable deep learning models to adaptively 
integrate various data modalities within a unified framework. In contrast to traditional EO models, 
which are typically specialized for specific modalities (optical, radar, or hyperspectral), 
DOFA leverages a dynamic hypernetwork architecture that adjusts its internal parameters based on 
the spectral characteristics of the input, specifically the central wavelengths. 
At its core, DOFA features a shared Vision Transformer backbone and a wavelength-conditioned dynamic patch embedding module, 
which allows it to handle inputs with varying numbers of channels from a wide range of sensors. 
This flexible design enables the model to dynamically generate weights and biases tailored to each modality, 
fostering effective cross-modal representation learning. The model was trained on a large-scale multimodal EO dataset 
of over 8 million images, curated from sources such as Sentinel-1, Sentinel-2, NAIP, Gaofen, and EnMAP, 
thus spanning SAR, RGB, multispectral, and hyperspectral domains. DOFA was evaluated on 14 downstream tasks 
spanning classification and segmentation, including those in the GEO-Bench benchmark suite. 
Remarkably, it outperformed or matched other latest pretrained models such as GFM, SatMAE, Scale-MAE, and SpectralGPT 
in 13 out of the 14 tasks. In our study, we used the DOFA-large variant, referred to as \textit{Dofa\_300M}, which contains approximately 330 million parameters.

\subsection{Ensembling of models}

Model ensembling is a widely used technique in machine learning that combines multiple models to enhance generalization, robustness, and overall performance 
across diverse tasks. However, this approach remains largely underexplored in the context of geospatial foundation models. 
Combining models such as Prithvi, which was trained on multispectral, medium-resolution time-series data, with SAM, 
which was trained primarily on high-resolution natural images and videos, is a promising way to leverage their complementary strengths 
across different modalities and spatial resolutions. If successful, ensembling foundation models could open up a new research direction 
that allows the development of powerful foundation models without the immense computational cost of training from scratch 
that currently is the dominant approach. Additionally, understanding which combinations of models work best may guide knowledge distillation strategies, 
enabling the creation of efficient models that retain the strengths of the ensemble while being more lightweight and cost-effective. 
In our study, we employed feature vector concatenation as the ensembling strategy. 
Specifically, the ensemble combined \textit{Prithvi\_300M} and \textit{Hiera\_200M}, and is referred to as \textit{Hiera\_Prithvi\_500M}.
This method retains the rich, high-dimensional representations of each individual model, 
although it comes with a higher computational cost compared to simpler techniques like averaging. 
Specifically, we extract the feature embeddings from each model and concatenate them before passing the combined representation into a lightweight, 
task-specific head, such as a segmentation decoder or classification module. Unlike output-level ensembling, which can suffer from modality mismatches, 
feature-level concatenation allows for deeper and more effective integration of multimodal information.
Due to limited computational resources, we were unable to explore additional models and combinations. 
This remains an area for future investigation.

\section{Experiments and Results}
\subsection{Experimental settings}

This study considers three downstream tasks: single-label classification, multi-label classification, and semantic segmentation. 
For each task, we designed a unified task-specific architecture into which all backbones can be seamlessly integrated. 
Specifically, for the classification tasks (both single-label and multi-label), 
we applied simple linear projection layers on top of the backbone's final layer output. 
These classification heads were adapted to each dataset to account for differences 
in the number of classes but were kept consistent across all backbones. 
For semantic segmentation, we employed the UPerNet architecture across all datasets \cite{Xiao2018}. 
While the network structure was slightly adapted to fit the specifics of each dataset, 
it was otherwise kept uniform across all backbone models. 
All implementations were efficiently carried out using the TerraTorch framework.

When using the GEO-Bench datasets, we followed the recommended evaluation protocol to ensure fair and reproducible model comparisons \cite{Lacoste2023,Szwarcman2024}. 
The procedure begins with hyperparameter tuning, where each model is given a fixed trial budget, 16 in our case, for each dataset. 
The best hyperparameters identified on the validation set are then used to run repeated experiments, 10 per dataset in our study, using different random seeds. 
This approach accounts for the inherent randomness in training AI models and enables more reliable performance comparisons.
Hyperparameter tuning was conducted using Bayesian Optimization via Optuna\footnote{https://optuna.readthedocs.io/en/stable/}. 
Across all experiments, the AdamW optimizer was used, and tuning was performed over a shared search space that included learning rate and weight decay. 
For comparability, the batch size was fixed to a reasonable value for each dataset and applied consistently across all models. 
Additionally, we restricted our experiments to optical sensor data, feeding into each backbone only the spectral bands on which it was pretrained. 
All input images were uniformly resized to 224 × 224 pixels across all datasets.

\subsection{Results and Discussion}

\begin{figure}[h]
  \centering
  
  \begin{subfigure}[b]{\textwidth}
    \centering
    \includegraphics[width=\linewidth]{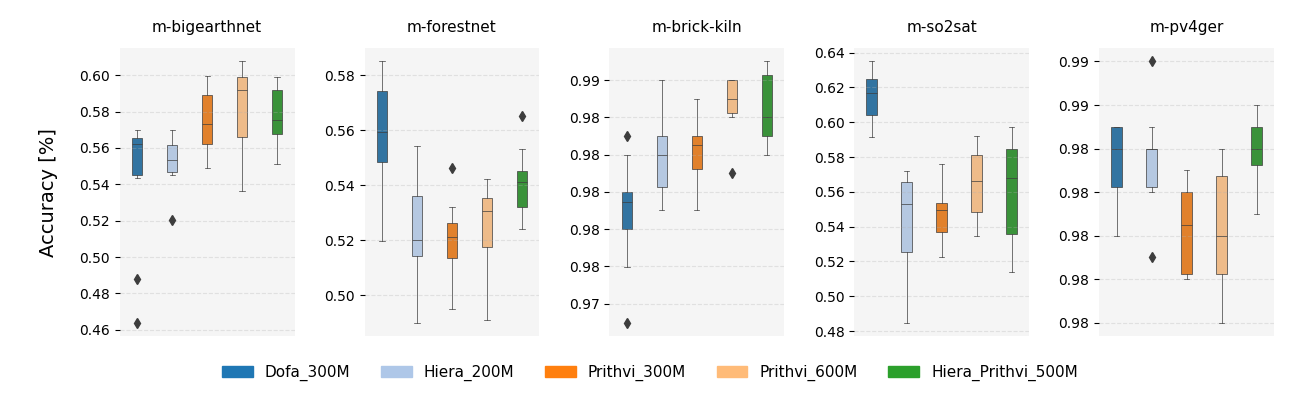}  % example image
    \caption{}
  \end{subfigure}
  
  \vspace{0.5cm}

  \begin{subfigure}[b]{\textwidth}
    \centering
    \includegraphics[width=\linewidth]{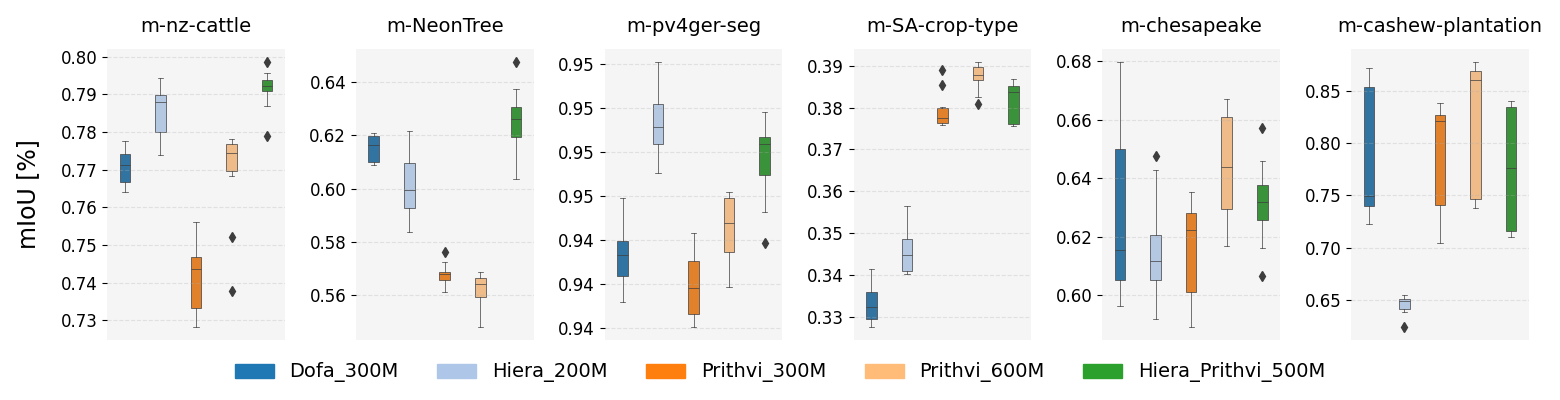}  % example image
    \caption{}
  \end{subfigure}

  \caption{Distribution of performance across all models from 10 repeated runs for the GEOBench (a) classification and (b) segmentation tasks in terms of accuracy and mean IoU,
respectively. We list input sensor and resolution for each dataset.}
  \label{fig:result}
\end{figure}

Figure~\ref{fig:result} presents the benchmark evaluation results across 11 remote sensing datasets using various foundation models.
The top row (a) shows classification accuracy, except for m-bigearthnet, a multilabel classification task for which the F1-score is used.
The bottom row (b) shows the mean Intersection over Union (mIoU) for semantic segmentation tasks.
Each boxplot summarizes performance over 10 independent runs with different random seeds to capture variability.
Overall, the results demonstrate that \textit{Hiera\_Prithvi\_500M} and \textit{Prithvi\_600M} are the top-performing models,
each achieving the best results on four datasets. They are followed by \textit{Dofa\_300M} and \textit{Hiera\_200M},
which achieved top performance on two datasets (m-forestnet, m-so2sat) and one dataset (m-pv4ger-seg), respectively.
Notably, despite using only three visible bands as input (see Table~\ref{tab:bandused}),
\textit{Hiera\_200M} outperforms larger models such as \textit{Prithvi\_300M} on seven tasks.

Feature-level ensembling, as demonstrated by \textit{Hiera\_Prithvi\_500M}, improves both accuracy and consistency across many cases.
The ensemble of \textit{Hiera\_200M} and \textit{Prithvi\_300M} consistently outperformed the individual models on 10 out of 11 datasets,
with the exception of the m-pv4ger-seg dataset. This highlights the potential of leveraging pretrained models 
through ensembling to achieve superior performance on key EO tasks.
\textit{DOFA} also emerged as a strong contender, outperforming models of similar size such as \textit{Prithvi\_300M} and \textit{Hiera\_200M} on six tasks.
This suggests that further exploration of model ensembling or knowledge distillation involving DOFA could be valuable in future work.
While larger models (e.g., \textit{Prithvi\_600M}) generally deliver stronger performance, they are significantly more resource-intensive.
Overall, the combined strengths of foundation models point toward a promising direction for geospatial AI.
The results support the notion that model ensembling can rival or even surpass large single models,
providing a computationally efficient alternative to training massive models from scratch.

\begin{table}[H]
	\centering
	\caption{Spectral bands fed into each model for each dataset. 
	Dataset names are shortened for better table formatting.
	RGB stands for Red, Green, and Blue.
	RGBN stands for Red, Green, Blue, and Near-Infrared (N).
	RGBNS1S2 stands for Red, Green, Blue, Near-Infrared, Shortwave Infrared 1 (S1), and Shortwave Infrared 2 (S2).}
	\label{tab:bandused}
	\resizebox{\textwidth}{!}{%
	\begin{tabular}{lccccccccccc}
		\toprule
		\textbf{Name} & \textbf{bigearthnet} & \textbf{so2sat} & \textbf{brick-kiln} & \textbf{forestnet} & \textbf{pv4ger} & \textbf{pv4ger-seg} & \textbf{chesapeake} & \textbf{cashew} & \textbf{SA-crop-type} & \textbf{nz-cattle} & \textbf{NeonTree} \\
		\midrule
		\textit{Hiera\_200M}          & RGB       & RGB       & RGB      & RGB      & RGB & RGB  & RGB  & RGB       & RGB      & RGB & RGB \\
		\textit{Dofa\_300M}           & RGBNS1S2  & RGBNS1S2  & RGBNS1S2 & RGBNS1S2 & RGB & RGB  & RGBN & RGBNS1S2  & RGBNS1S2 & RGB & RGB \\
		\textit{Prithvi\_300M}        & RGBNS1S2  & RGBNS1S2  & RGBNS1S2 & RGBNS1S2 & RGB & RGB  & RGBN & RGBNS1S2  & RGBNS1S2 & RGB & RGB \\
		\textit{Prithvi\_600M}        & RGBNS1S2  & RGBNS1S2  & RGBNS1S2 & RGBNS1S2 & RGB & RGB  & RGBN & RGBNS1S2  & RGBNS1S2 & RGB & RGB \\
		\textit{Hiera\_Prithvi\_500M} & RGBNS1S2  & RGBNS1S2  & RGBNS1S2 & RGBNS1S2 & RGB & RGB  & RGBN & RGBNS1S2  & RGBNS1S2 & RGB & RGB \\
		\bottomrule
	\end{tabular}%
	}
\end{table}

Additional analysis suggests that Hiera-based backbones may benefit from larger original input sizes. 
Both \textit{Hiera\_200M} and \textit{Hiera\_Prithvi\_500M} generally outperformed other backbones on datasets 
with larger image dimensions, such as m-pv4ger (320×320), m-NeonTree (400×400), and m-nz-cattle (500×500). 
Furthermore, experiments using different random seeds showed lower performance variance on these datasets compared to those with smaller input sizes.
In addition, Hiera-based models tend to perform better on high-resolution datasets, 
including those with 0.1-meter resolution (m-pv4ger, m-pv4ger-seg, m-nz-cattle, and m-NeonTree) 
and 1-meter resolution (m-chesapeake-landcover). This behavior may be attributed to 
the original pretraining resolution of the Hiera models, which were trained on 1024×1024 images, 
primarily consisting of natural camera imagery. This characteristic of the Hiera backbone 
is important to consider in future applications, particularly when preparing datasets for finetuning or other downstream tasks.

\section{Conclusion}

In this study, we explored a novel direction in the development of geospatial foundation models 
by leveraging existing pretrained models to achieve state-of-the-art performance. 
We conducted a comprehensive benchmark evaluation of recent representative vision foundation models 
from both the geospatial and general computer vision domains. Additionally, we evaluated feature-level ensembling 
of two representative models, namely Hiera and Prithvi. The models were assessed on a diverse set of Earth observation tasks using 
the GEO-Bench framework. Results demonstrate that recent models such as \textit{Prithvi\_600M} achieve top performance 
in many classification and segmentation tasks. Notably, feature-level ensembling of smaller models like \textit{Hiera\_200M} and \textit{Prithvi\_300M} 
proved highly effective, achieving competitive results on multiple datasets, rivaling the larger, siloed model. 
This highlighs the potential of combining pretrained models for improved accuracy and robustness.

Our analysis further reveals that Hiera-based models exhibit stronger performance on datasets 
with higher spatial resolutions and larger input sizes, likely due to their pretraining on 1024×1024 natural images. 
This insight has important implications for downstream applications, suggesting that aligning the characteristics 
of target datasets with the model’s pretraining configuration could enhance transfer learning effectiveness.

While larger models such as \textit{Prithvi\_600M} deliver strong performance, 
they require substantial computational resources. Our findings suggest that model ensembling 
provides a promising and more resource-efficient alternative, capable of matching 
or even surpassing the performance of these larger individual models. 
This highlights the importance of flexible and modular model design in advancing geospatial AI. 
Future research may focus on exploring knowledge distillation techniques 
to compress ensembles into lightweight models suitable for real-world deployment.

\section*{Acknowledgement}

This work was supported by the International Digital Earth Applied Science Research Center, 
an International Joint Usage/Research Center at Chubu University, Japan.
We gratefully acknowledge Dr. Tran Van Hien of the National Institute of Information and Communications Technology, Japan, 
for his invaluable support in training the models used in this study.

% \nocite{*}  % Include all entries in .bib
% \bibliography{references}

\end{document}